\pdfoutput=1

\documentclass[11pt]{article}

\usepackage{EMNLP2023}

\usepackage{times}
\usepackage{latexsym}

\usepackage{graphicx}
\usepackage[T1]{fontenc}

\usepackage[utf8]{inputenc}

\usepackage{microtype}

\usepackage{inconsolata}

\title{NeuroSurgeon: A Toolkit for Subnetwork Analysis}

\author{Michael A. Lepori \\
Department of Computer Science \\
  Brown University \\
  \texttt{michael\_lepori@brown.edu} \\\And
  Ellie Pavlick \\
Department of Computer Science \\
  Brown University \\
  \texttt{ellie\_pavlick@brown.edu} \\ \AND
  Thomas Serre \\
  Carney Institute for Brain Science\\
  Brown University \\
  \texttt{thomas\_serre@brown.edu}
}
\begin{document}

\maketitle
\begin{abstract}
Despite recent advances in the field of explainability, much remains unknown about the algorithms that neural networks learn to represent. Recent work has attempted to understand trained models by decomposing them into functional circuits \cite{csordas2020neural, lepori2023break}. To advance this research, we developed \textbf{NeuroSurgeon}, a python library that can be used to discover and manipulate subnetworks within models in the Huggingface Transformers library \cite{wolf2019huggingface}. NeuroSurgeon is freely available at \url{https://github.com/mlepori1/NeuroSurgeon}.

\end{abstract}

\section{Introduction}
Neural networks -- particularly transformers \cite{vaswani2017attention} -- are the de facto solution to machine learning problems in both industry and academia.
Despite their ubiquity, these models are largely inscrutable. Recent work in mechanistic interpretability has manually reverse-engineered specialized circuits\footnote{In this work, we define a ``circuit'' as a portion of a model that performs some high-level functions, and a ``subnetwork'' as any subset of weights or neurons within a model. A circuit can thus be localized to a subnetwork, and a subnetwork can comprise a circuit if it performs a high-level function.} in small models, but scaling up this approach poses a daunting challenge \cite{nanda2022progress, merullo2023language, wang2022interpretability, olsson2022context}. 

Another line of work employs \textit{subnetwork analysis} to understand the internal structure of trained models. This approach seeks to automatically uncover circuits within a trained model and locate them in particular subnetworks. This approach borrows techniques from model pruning to uncover subnetworks that might implement such high-level computations. We developed  a python library -- \textbf{NeuroSurgeon} -- to simplify the process of subnetwork analysis, allowing researchers to more quickly uncover the internal structure that lies within trained models. 

\section{Overview}
  NeuroSurgeon supports several popular models within the Huggingface Transformers repository \cite{wolf2019huggingface}, including ViT \cite{dosovitskiy2020image}, ResNet \cite{he2016deep}, GPT2 \cite{radford2019language}, BERT \cite{devlin2018bert}, and more. 
  With NeuroSurgeon, one discovers functional subnetworks by optimizing a binary mask over weights (or neurons) within model layers, ablating everything except the units necessary for a particular computation.
We have implemented two optimization-based techniques from model pruning (as well as a simple baseline technique) for generating these binary masks.

\paragraph{Hard-Concrete Masking:}
Hard-Concrete masking was introduced to provide an approximation to the $l_0$ penalty, providing a bias towards sparse solutions during model training \cite{louizos2017learning}. This technique produces masks by stochastically sampling mask values from a parameterized hard-concrete distribution. 
\paragraph{Continuous Sparsification:}
Continuous Sparsification was introduced to provide a \textit{deterministic} approximation to the $l_0$ penalty \cite{savarese2020winning}. This technique produces masks by annealing a parameterized soft mask into a hard mask over the course of training. 

\paragraph{Magnitude Pruning:}
Magnitude pruning simply ablates some fraction of the lowest magnitude weights \cite{han2015learning}. Though simple, this approach has been used in several important works on pruning and subnetworks, notably the Lottery Ticket Hypothesis \cite{frankle2018lottery}. This method should be used as a baseline to compare against the optimization-based methods described above.

\paragraph{}When performing subnetwork analysis, we freeze the underlying model weights and optimize the parameters introduced by Continuous Sparsification or Hard-Concrete Masking. We typically include an $l_0$ regularization term on the mask to encourage parsimonious subnetworks. Both optimization-based techniques can be used to discover subnetworks at the weight or neuron level.

\section{Visualization}
\begin{figure}
\centering
\includegraphics[width=\linewidth]{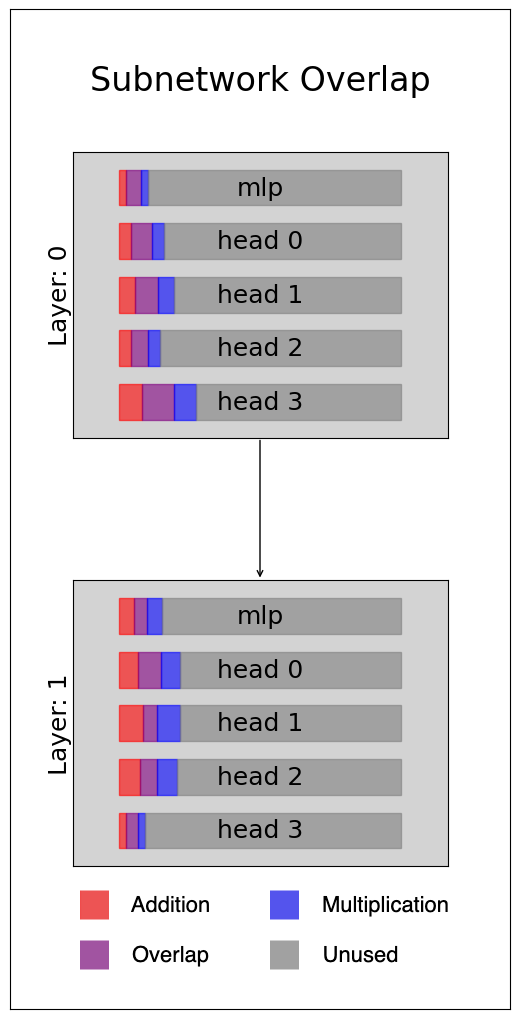}
\caption{Visualization of two subnetworks within a 2-layer GPT2-style transformer. This transformer was trained in a multitask fashion on addition and multiplication tasks, similar to the Addition/Multiplication setting in \citet{csordas2020neural}. One subnetwork was optimized to solve addition problems and the other was optimized to solve multiplication problems. Both were trained with $l_0$ regularization. Notably, we see that the subnetworks are sparse -- the majority of each block was pruned. Additionally, we see more subnetwork overlap in Layer 0 than in Layer 1. For instance, the subnetworks are almost entirely overlapping in Layer 0's MLP. On the other hand, Layer 1's MLP and Attention Heads 1 and 2 contain very little overlap.}
\label{viz_fig}
\end{figure}

In order to visualize the results of subnetwork analysis, we have implemented a visualizer that can be used to understand how subnetworks are distributed throughout the layers of a model. It can be used to display one or two subnetworks within the same model. See Figure~\ref{viz_fig} for an example visualization of two subnetworks in a 2-layer GPT2-style transformer.

\section{Related Work}
Subnetwork analysis has been used in a wide variety of contexts in recent deep learning research. Some studies have used subnetwork analysis to uncover how linguistic information is distributed throughout a model \cite{de2022sparse, de2020decisions}. One notable approach to this is \textit{subnetwork probing} \cite{cao2021low}, which NeuroSurgeon implements. Others have sought to understand how particular computations are structured within model weights \cite{csordas2020neural, lepori2023break, conmy2023towards}. Still others have used subnetwork analysis to better understand generalization and transfer learning \cite{zhang2021can, panigrahi2023task, zheng2023regularized, guo2021parameter}, or to control model behavior \cite{li2023circuit}.

\section{Discussion}
We present NeuroSurgeon, a python library designed to enable researchers to easily identify functional subnetworks within trained models. We hope that NeuroSurgeon lowers the barrier to entry for researchers interested in performing subnetwork analysis for mechanistic interpretability.
\bibliography{emnlp2023}
\bibliographystyle{acl_natbib}

\end{document}